# Classification of Single and Mixed Partial Discharges under Switching Voltage Using an AWA–CNN Framework

Md Rafid Kaysar Shagor, *Student Member, IEEE,* Zannatul Ferdousy Mouri, Student *Member, IEEE,* Farhina Haque, *Member, IEEE,* Anindya Bijoy Das, *Member, IEEE*

***Abstract*— The growing use of fast-switching power electronics has made partial discharge (PD) analysis under switching-voltage excitation increasingly important, yet more challenging than under sinusoidal conditions due to activity concentrated at voltage transitions. This work presents an Amplitude-Width-Area (AWA) pattern representation for source-oriented PD analysis under switching-voltage excitation. In the proposed method, time-domain PD pulses are characterized using pulse amplitude, width, and area, and mapped into a visual pattern where amplitude and area define the coordinate axes and width is encoded by color. The generated AWA patterns are used to distinguish six single and mixed PD source conditions: corona, internal, surface, corona+internal, corona+surface, and internal+surface. To evaluate the classification capability of the proposed representation, a Random Forest baseline and two Convolutional Neural Network (CNN) models, InceptionV3 and ResNet-18, are compared. The AWA patterns show distinguishable source-dependent distributions, and CNN-based classification achieves testing accuracy above 96%, compared with 73.33% for Random Forest. The results indicate that AWA patterns provide a visual representation of PD pulses suitable for multi-class PD source classification under switching-voltage excitation.**

## I. Introduction

THE rapid advancement of power electronic (PE) systems, driven by the adoption of wide bandgap (WBG) semiconductor devices, has enabled significant improvements in modern power conversion, including higher switching speeds, enhanced efficiency, and more compact system designs.[1], [2]. Owing to these advantages, they are being increasingly deployed in electric vehicles, renewable energy conversion systems, more-electric aircraft, and shipboard power networks [3], [4], [5], [6]. However, the insulation system in PE-driven systems, including motor windings, power cables, busbars, and power module substrates, are exposed to fast-switching electrical stress during operation [7]. Under high-dv/dt repetitive switching voltage, insulation materials experience localized electric-field enhancement and accelerated aging [8]. As a result, insulation reliability has become an important concern for the safe and stable operation of modern PE-driven equipment. In this context, partial discharge (PD), which arises from local electric field enhancement in insulation, serves as an indicator of early ageing of insulation. Therefore, PD monitoring plays an important role in detecting early ageing, insulation assessment, and improving the long-term reliability of switching-voltage-based PE systems [9], [10].

PD activity is typically classified into three main types: corona discharge caused by sharp edges in air, internal discharge occurring within cavities, and surface discharge at triple-point regions. Each discharge mechanism shows different degradation behavior and multiple discharge mechanisms can occur simultaneously [11], [12]. For this reason, PD source identification is important for the dielectrically robust operation of PE-driven systems. Under conventional sinusoidal AC excitation, phase-resolved partial discharge (PRPD) analysis has been widely used for this purpose. In PRPD analysis, discharge magnitude and occurrence are mapped with respect to the phase angle of the applied voltage, producing characteristic patterns that help distinguish different PD types [11], [13]. For this purpose, Convolutional Neural Network (CNN)-based models have been used to learn source-dependent features from PRPD images and automate PD source classification [14]. However, in PE systems driven by switching or pulse-type voltages, the excitation waveform is no longer sinusoidal, and PD activity tends to occur near fast voltage transition regions rather than being distributed over a complete phase cycle [15]. As a result, conventional PRPD-based interpretation becomes less effective for switching-voltage PD analysis, motivating the need for alternative pattern representations that can describe PD behavior directly from pulse-domain measurements [16].

This paragraph of the first footnote will contain the date on which you submitted your paper for review, which is populated by IEEE. It is IEEE style to display support information, including sponsor and financial support acknowledgment, here and not in an acknowledgment section at the end of the article. For example, "This work was supported in part by the U.S. Department of Commerce under Grant 123456." The name of the corresponding author appears after the financial information, e.g. *(Corresponding author: M. Smith)*. Here you may also indicate if authors contributed equally or if there are co-first authors.

Md Rafid Kaysar Shagor, Zannatul Ferdousy Mouri, Farhina Haque, and Anindya Bijoy Das are with the Department of Electrical and Computer Engineering, The University of Akron, Akron, OH 44325 USA (e-mail: rs467@uakron.edu; zm66@uakron.edu; fhaque@uakron.edu; adas@uakron.edu).

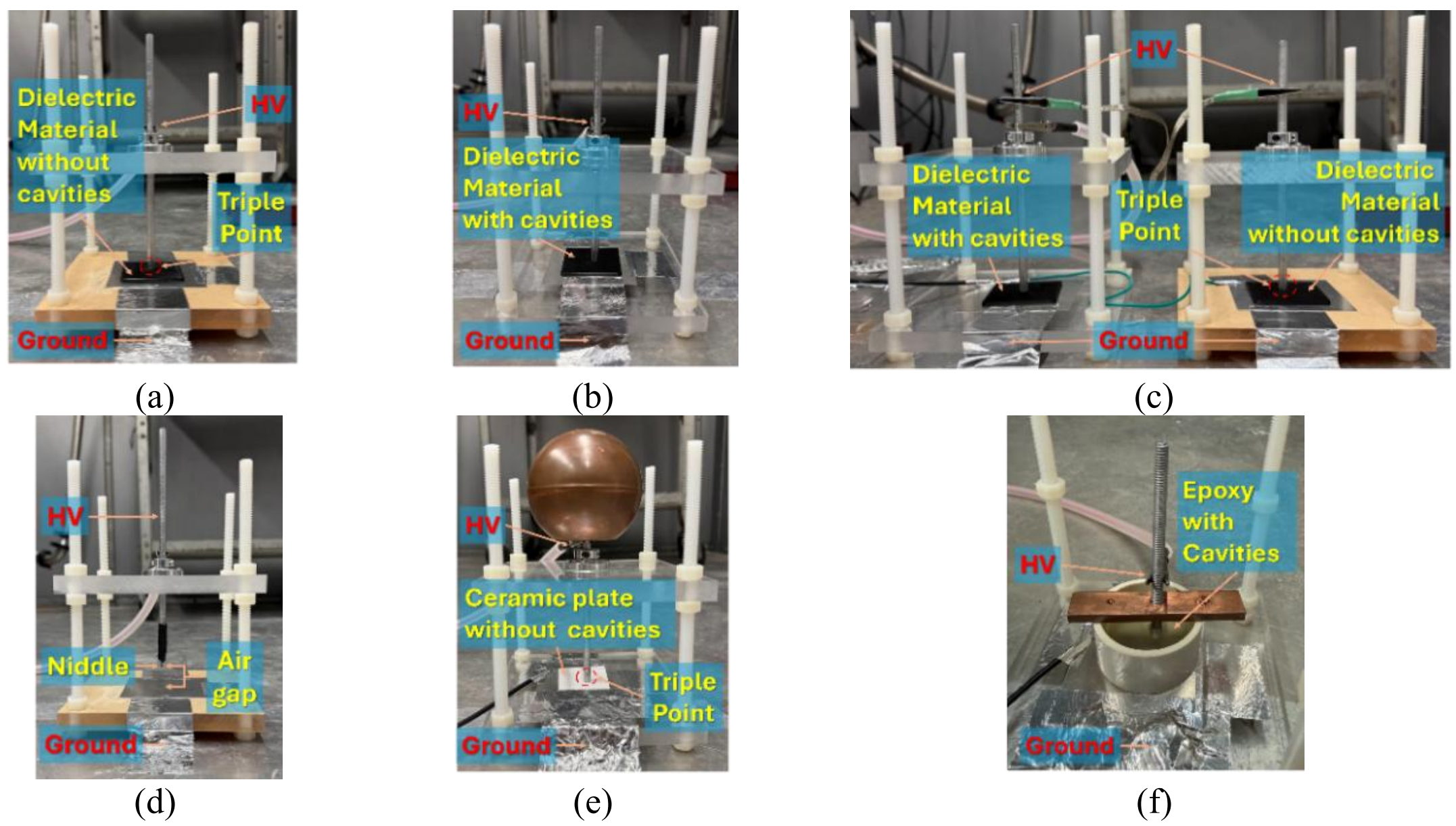

Fig. 1. Fabricated PD test samples and verification samples for (a) surface PD, (b) internal PD, (c) mixed PD, (d) corona PD, (e) ceramic-plate surface-PD verification sample, and (f) cured epoxy internal-PD verification sample.

Numerous studies have been conducted to investigate the PD behavior under PWM and pulse-type excitation. Under repetitive PWM pulses, Guo et al. reported that PD activity is mainly associated with the rising and falling edges of the waveform, rather than being distributed throughout the voltage cycle as in conventional phase-resolved patterns [15]. Jiang et al. further investigated pulse-voltage PD behavior using time-resolved measurements and showed that discharge events form narrow clusters around switching instants, with the distribution affected by excitation frequency [17]. These findings show that PD under switching voltages is concentrated around pulse transitions and cannot be fully interpreted using assumptions developed for sinusoidal excitation. Balouji et al. demonstrated that PD signals for different cavities under multilevel PWM excitation contain classifiable information by applying machine learning to pulse-domain features such as maximum amplitude, duration, inter-pulse interval, and waveform area [18]. However, existing approaches largely focus on temporal analysis, PRPD or time-density visualization, or manually engineered features. Therefore, a phase-independent, pulse-derived pattern representation is needed to translate switching-transition PD characteristics into a form suitable for PD source classification.

To address this limitation, this work presents an Amplitude–Width–Area (AWA)-based pattern generation approach for PD source classification under switching-voltage excitation involving both single-source and mixed-source PD mechanisms. In the proposed approach, individual PD pulses measured in the time domain are characterized using three pulse-level quantities: amplitude, width, and area, where the area is obtained from the product of pulse amplitude and width. These features are mapped into a two-dimensional visual representation, where amplitude and area form the coordinate axes and width is encoded through color. The resulting AWA images form the input dataset for six-class PD source classification, including corona (C), internal (I), surface (S), corona + internal (CI), corona + surface (CS), and surface + internal (SI) discharges. As the proposed representation transforms pulse-domain data into image-like feature distributions, CNN-based transfer learning is employed to capture spatial relationships and assess class separability. InceptionV3 and ResNet-18 are used and compared for AWA pattern classification. A Random Forest classifier is also included as a traditional machine learning (ML) baseline to assess the benefit of CNN-based pattern learning for AWA images. The evaluation indicates that the proposed AWA–CNN framework provides an effective approach for multi-class PD source classification under switching-voltage conditions.

## II. Switching-Voltage PD Measurement and Source Configuration

### *A. PD Source Configuration for Single and Mixed Discharges*

PD source configuration was designed to generate both single-source and mixed-source discharge conditions under switching-voltage excitation, as shown in Fig. 1. Three single PD sources were considered: corona, internal, and surface discharge. The surface PD sample (Fig. 1(a)) was fabricated using a 3D-printed PLA dielectric with 100% infill, promoting discharge along the surface near the electrode–dielectric–air interface. The internal-PD sample (Fig. 1(b)) was prepared using a 3D-printed PLA dielectric with intentionally introduced void-like air regions which promoted discharge activity inside the material. Corona discharge (Fig. 1(d)) was generated using a needle-plane configuration where the highly non-uniform electric field near the needle initiates discharge in the surrounding air. To represent insulation conditions where multiple defects coexist, three mixed PD configurations were prepared: corona + internal, corona + surface, and surface +internal. These mixed-source conditions were produced by connecting the corresponding single-source configurations in parallel within the same measurement arrangement for simultaneous PD activities (Fig. 1(c)). In addition, ceramic and cured epoxy with cavity samples were prepared for verification of AWA pattern consistency (Fig. 1(e) and Fig. 1(f)).

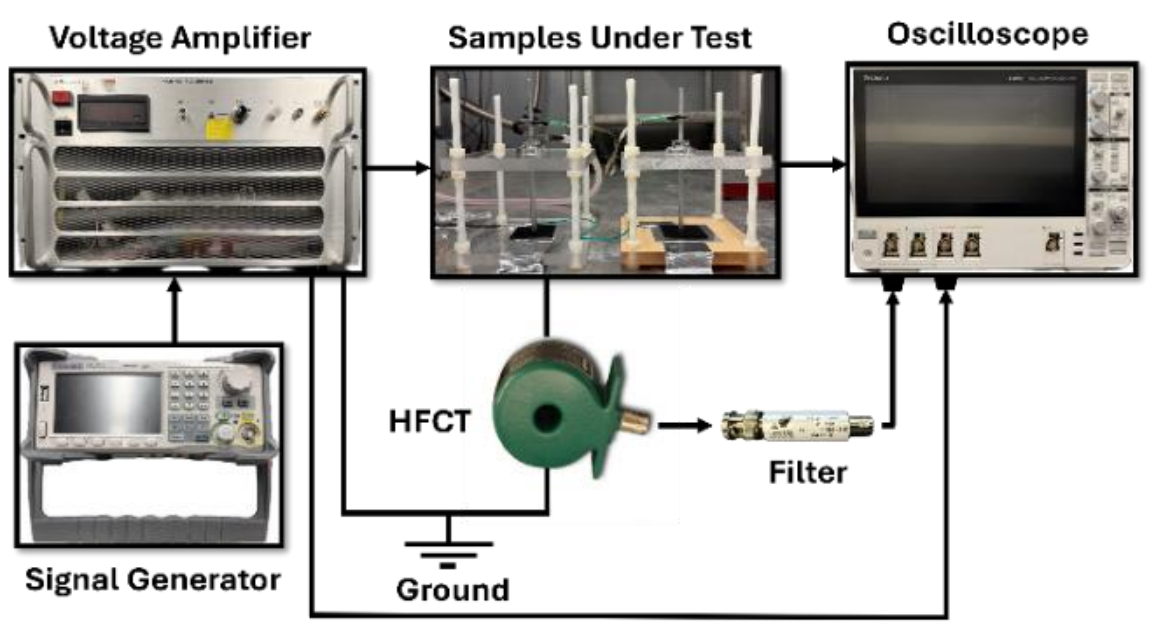

Fig. 2. Experimental setup for PD measurement under power electronic switching voltage based on IEC 60270 [19].

### B. Experimental Setup and Switching-Voltage Excitation

The PD experiments were carried out using an arrangement designed, as shown in Fig. 2, to apply controlled switching voltage stress to the prepared samples. A pulse waveform with a frequency of 60 Hz, a duty cycle of 50%, and rise and fall times of 18 µs was generated using a signal generator and then amplified by a high-voltage amplifier with a gain of approximately 2000 before being applied to the test sample configuration. These waveform parameters were selected to provide repeatable switching-voltage excitation and generate PD in various source configurations under sufficient electrical stress. For each sample, the voltage amplitude was gradually increased until PD activity was observed, and the corresponding partial discharge inception voltage (PDIV) was recorded [20]. The measured PDIV values were approximately 7.88 kVp for surface PD, 7.34 kVp for internal PD, and 8.2 kVp for corona PD. For mixed-source configurations, the applied voltage was selected to enable simultaneous discharge activity from the corresponding paired sources. This procedure allowed the corona, internal, surface, and mixed PD configurations to be investigated under a consistent measurement framework.

### C. PD Measurement and Signal Acquisition

PD signals were measured using a high-frequency current transformer (HFCT, 300 Hz – 200 MHz) placed in the return path of the test circuit. The HFCT was used to capture the high-frequency current pulses associated with PD activity under switching-voltage excitation. The PD signal detected by the HFCT was passed through a 50 MHz high-pass filter to suppress switching-induced noise and then recorded using a high-frequency oscilloscope. Fig. 3 shows the applied switching voltage and the corresponding PD signal for consecutive 20 cycles captured by the HFCT. It is clear from the figure that PD occurs at the rising and falling edges of the pulse excitation, as mentioned in the previous literatures [15], [17]. For further study, time and amplitude information of the time-domain PD signals were saved separately as CSV files. These recorded signals provided the raw data foundation for the offline feature extraction, AWA pattern generation, and classification framework described in the following sections.

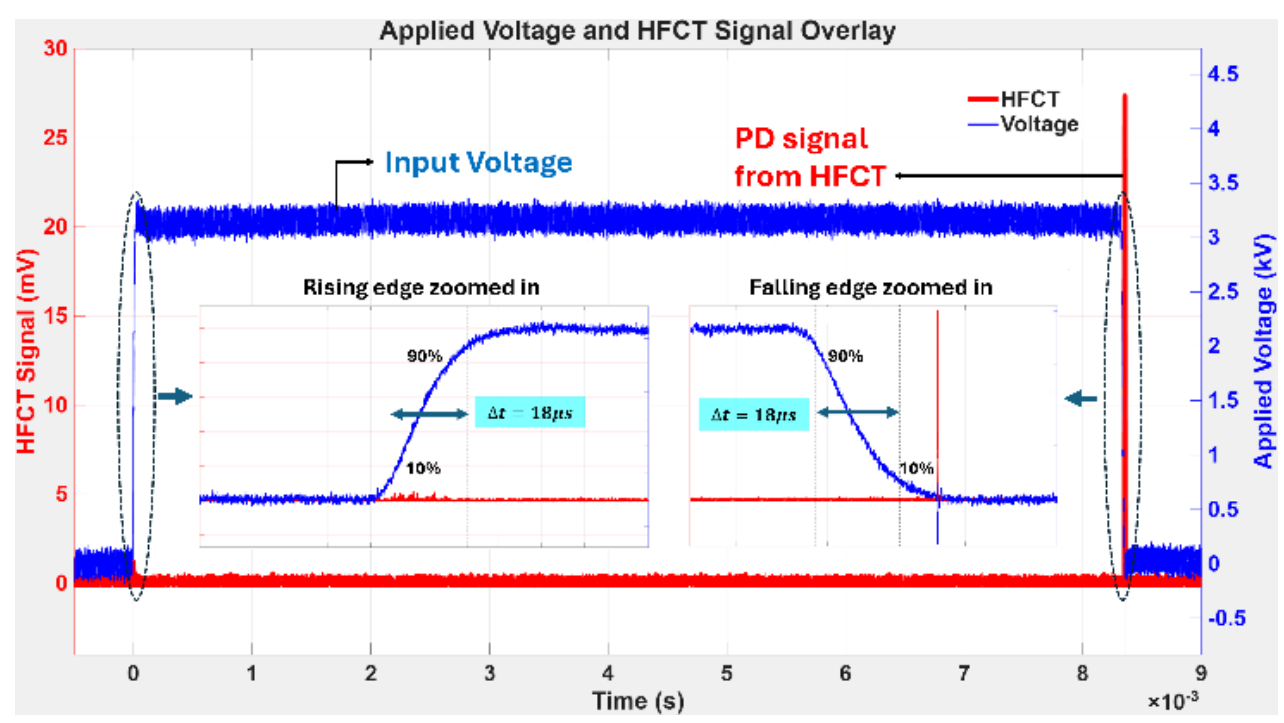

Fig. 3. Applied switching voltage and corresponding PD signal captured by the HFCT.

## III. AWA-Based PD Pattern Representation

### A. AWA Feature Definition and Pattern Construction

The AWA-based representation was constructed from time-domain PD signals obtained from stored CSVs for each source condition. Since the acquired CSV files contain time and amplitude information, individual PD pulses were identified using a peak-detection procedure at the switching edge. In this work, the MATLAB findpeaks function was used to detect local maxima and return associated peak properties, including peak height, location, width, and prominence [21], [22]. In the peak-selection process, amplitude-based and prominence-based criteria were used together, where amplitude criterion was used to suppress low-level background noise, and prominence criterion ensured a locally distinguishable peak from the surrounding waveform. In this context, prominence describes how much a peak rises relative to its neighboring baseline, making it useful for separating PD-related pulses from small fluctuations or residual switching-related disturbances. For each accepted PD pulse, the peak height was defined as the pulse amplitude, the peak width measured at the half-prominence level was used as the pulse width, and the pulse area was calculated as the product of amplitude and width. The width was measured at the half-prominence level because it provides a consistent local reference for pulse-width extraction while reducing the influence of low-amplitude tails and local waveform fluctuations [23]. The relationship among these extracted quantities is illustrated in Fig. 4.

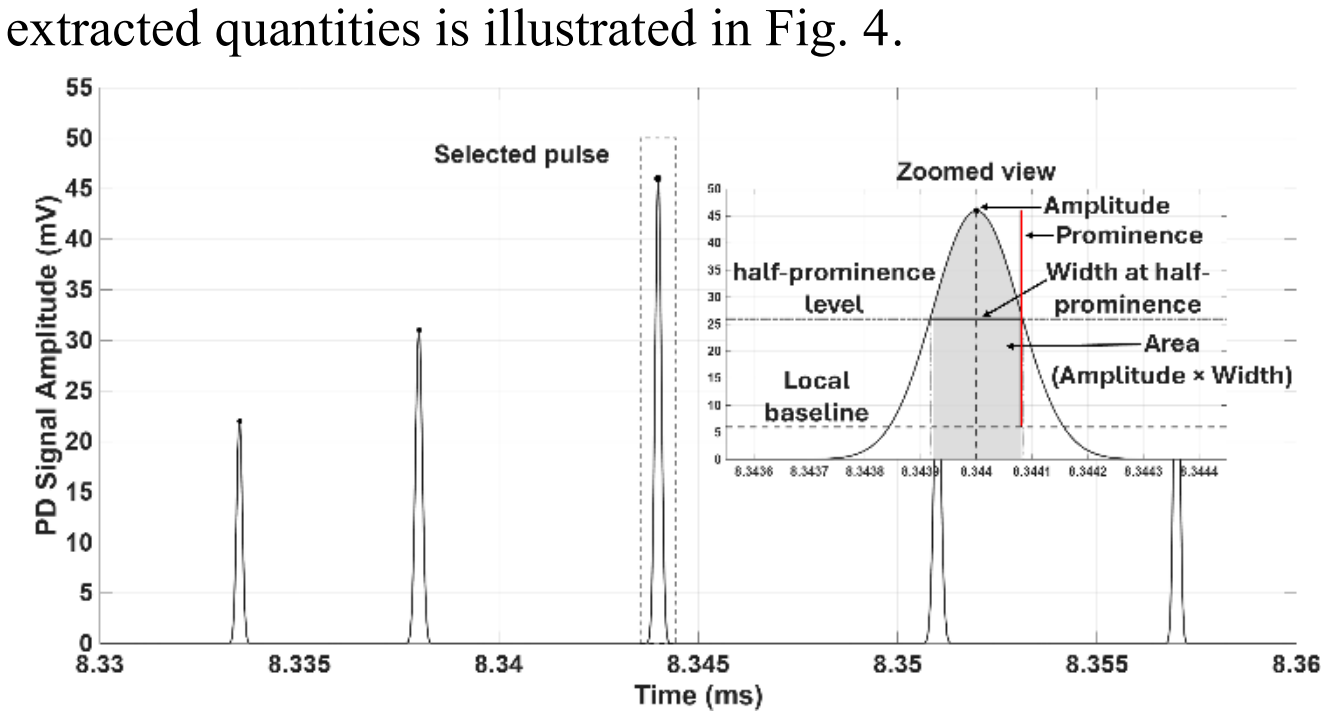

Fig. 4. Simplified representation of a detected PD pulse showing the extracted AWA features: amplitude, width, and area.

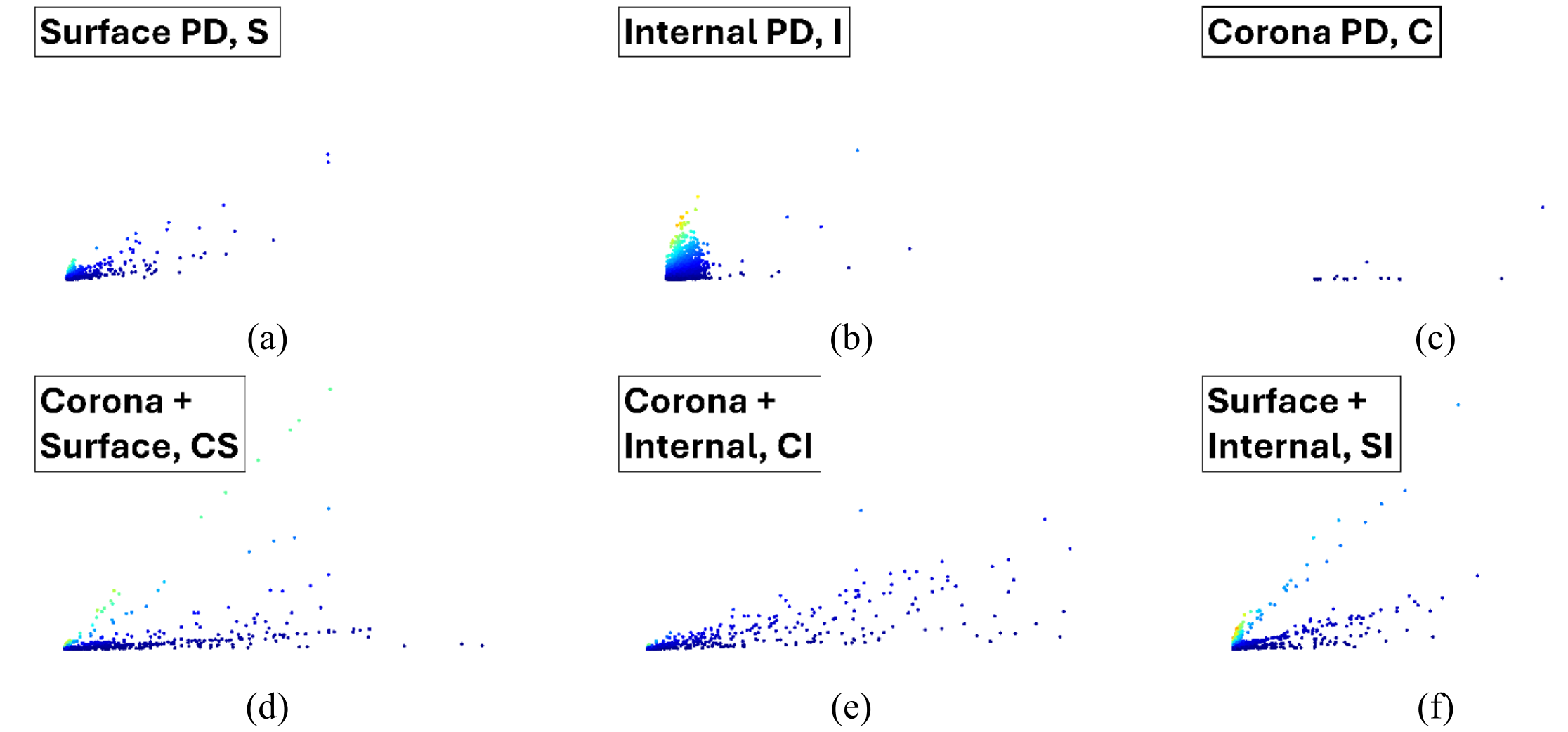

Fig. 5. AWA-based PD patterns for (a) surface (S), (b) internal (I), (c) corona (C), (d) corona + surface (CS), (e) corona + internal (CI), and (f) surface + internal (SI) PD types.

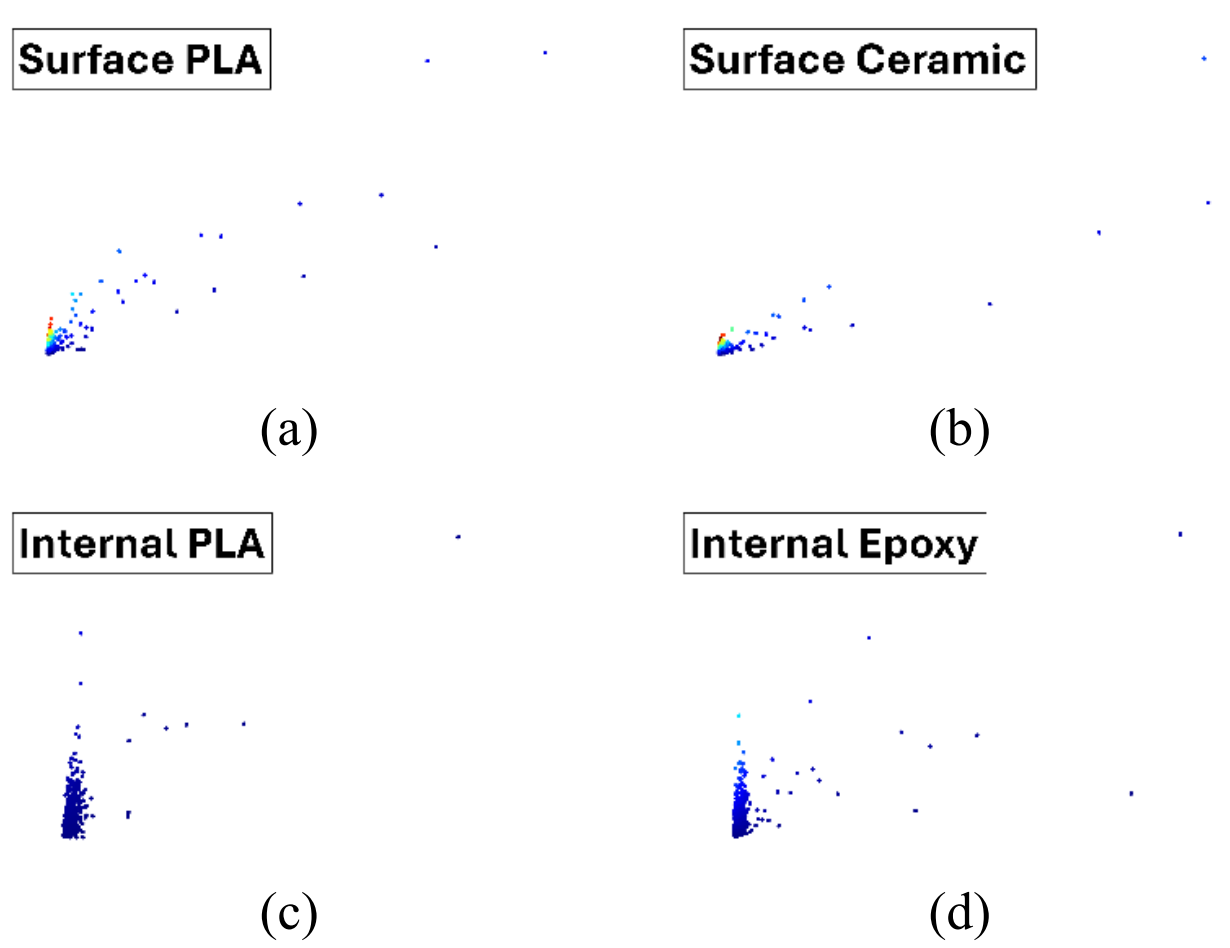

Fig. 6. Material-independent AWA patterns, (a) surface PD using PLA, (b) surface PD using ceramic plate, (c) internal PD using PLA with cavities, and (d) internal PD using epoxy with air bubbles.

After feature extraction, each PD pulse was represented using its amplitude, width, and area values. These quantities were then mapped into a two-dimensional AWA pattern where amplitude was assigned to the horizontal axis and area was assigned to the vertical axis. The pulse width was encoded using color intensity, allowing each discharge event to be represented as a colored point in the amplitude–area feature space. In this way, the AWA pattern combines information related to discharge magnitude, duration, and energy-related behavior within a single visual representation.

### B. AWA Pattern Characteristics of Single and Mixed PD Sources

The AWA patterns corresponding to various PD source conditions exhibit distinct distribution characteristics in the amplitude-area feature space, as shown in Fig. 5. For single-source PDs, surface PD (Fig. 5(a)) shows a dense cluster near the lower amplitude and area region, accompanied by a pronounced fan-shaped distribution extending toward higher amplitude and area values. Internal PD (Fig. 5(b)) also shows a dense cluster near the lower amplitude and area region but is distinguished by a vertically elongated distribution along the area axis, with relatively limited spread in amplitude. In contrast, corona PD (Fig. 5(c)) exhibits a sparse distribution with a limited number of points concentrated within a narrow amplitude-area region and occasional isolated points at higher amplitude levels. For mixed-source conditions, the AWA patterns reflect combined characteristics of the corresponding single-source distributions while maintaining distinguishable structural features. The CS (Fig. 5(d)) pattern exhibits a strong concentration near the origin, similar to corona PD, together with a fan-shaped extension toward higher amplitude and area values associated with surface discharge behavior. The CI (Fig. 5(e)) pattern shows a dense low-area region with a noticeable horizontal spread along the amplitude axis and a relatively limited extension in the area direction. In contrast, the SI pattern (Fig. 5(f)) presents a broader and more structured distribution, combining the fan-shaped spread of surface PD with the vertical extension associated with internal PD. Overall, the AWA representation produces visually distinct distributions for different PD source conditions while retaining characteristic features of both individual and combined discharge behaviors. These observable differences provide a reliable basis for distinguishing multiple PD source types and support the use of learning-based classification models under switching-voltage excitation.

### C. Verification of AWA Pattern Consistency Across Different Materials

To verify that the AWA patterns for each defect class are independent of materials properties, additional surface and internal PD measurements were performed using ceramic and epoxy-based samples. The surface and internal PD patterns discussed in III (B) were generated using 3D-printed PLA samples. For surface PD, the PLA sample without cavities was compared with the alumina ceramic-plate configuration shown in Fig. 1(e). In both cases, discharge activity was initiated at the triple-junction region formed by the electrode, dielectric surface, and surrounding air under switching-voltage excitation [24]. As shown in Fig. 6(a) and Fig. 6(b), the corresponding AWA patterns exhibit similar surface-discharge characteristics, with a broader spread along the amplitude direction.

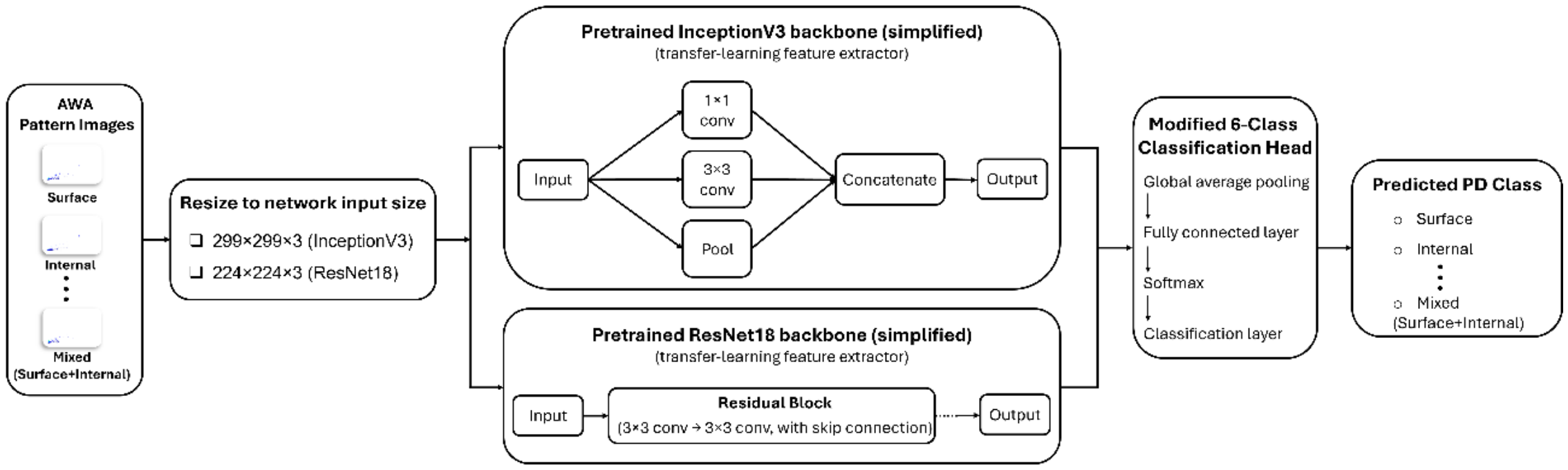

Fig. 7. Simplified transfer-learning workflow for PD classification using InceptionV3 and ResNet-18.

Similarly, the internal PD pattern obtained from the 3D-printed PLA sample with internal cavities was compared with the cured epoxy sample containing embedded air bubbles produced by inserting a needle during curing of epoxy, as shown in Fig. 1(f). These air-filled regions served as internal discharge sources under sufficiently high switching-voltage stress [25]. The AWA patterns shown in Fig. 6(c) and Fig. 6(d) exhibit similar internal-discharge characteristics where the discharge points remain more concentrated in the lower-amplitude region compared with surface PD. These observations indicate that the AWA patterns achieved with various sources are independent of material properties. Therefore, the verification results support the use of AWA patterns as a source-oriented visual representation for PD classification under switching-voltage excitation.

## IV. AWA–CNN Framework for PD Classification

### A. AWA Pattern Dataset Preparation

The AWA patterns generated from the extracted PD pulse features were organized as image datasets for classification. Each AWA image was labeled according to one of the six PD source classes: C, I, S, CI, CS, and SI as shown in Fig. 5. The original dataset contained 250 AWA images for each PD source class. To reduce the possibility of data leakage, the images in each class were first divided into training, validation, and testing subsets using a 60:20:20 split, corresponding to 150, 50, and 50 images per class, respectively. Data augmentation was then performed separately within the training, validation, and testing subsets to ensure that augmented versions derived from the same AWA image did not appear in any other subset. The augmentation methods included scaling, Gaussian blurring, brightness and contrast adjustment, shearing, rotation, and horizontal flipping [26]. These transformations were used to expand the dataset under controlled image variations while preserving the overall AWA distribution characteristics of each PD source. After augmentation, each class contained 750 training images, 250 validation images, and 250 testing images, resulting in 1250 images per class and 7500 images in total across the six PD source classes. A file-level integrity check was also performed after dataset preparation, and no overlapping image files were found among the training, validation, and testing folders. Before classification, the AWA images were formatted according to the input requirements of each model, e.g., for the Random Forest baseline, the images were converted into handcrafted numerical feature vectors, whereas for the CNN-based models, the images were resized according to the corresponding network input dimensions.

### B. Traditional Machine Learning Feature Extraction and Training Procedure

Traditional ML methods are widely used in classification tasks where input data or images are represented by extracted numerical features before classifier training [27]. For image-based classification, these features may summarize statistical, spatial, color, or texture-related information. Random Forest is a traditional ML method that forms an ensemble of decision trees trained using random subsets of samples and features, with the final prediction obtained from the combined tree decisions [28], [29]. This structure allows Random Forest to model nonlinear feature relationships in multi-class classification problems.

In this work, Random Forest was used as a conventional ML baseline to evaluate handcrafted features derived from the AWA pattern images and to compare model simplicity, feature interpretability, implementation complexity, and computational demand. Each AWA image was converted into a numerical feature vector by resizing it to a fixed input size and separating the foreground AWA pattern from the image background. Shape-based and statistical features, including image kurtosis, foreground pixel count, foreground fraction, bounding-box dimensions, and aspect ratio, were extracted from the segmented AWA pattern. Color-based features were extracted from the red, green, and blue channels to represent the width-related color information embedded in the AWA pattern. In addition, the image was partitioned into grid cells, and the foreground occupancy in each cell was calculated to describe the local distribution of PD points across the amplitude–area feature space. These handcrafted features were then used to train the Random Forest classifier.

### C. CNN Architecture and Training Procedure

While traditional ML methods depend on predefined numerical features, convolutional neural networks (CNNs) are designed to learn representative features directly from images.

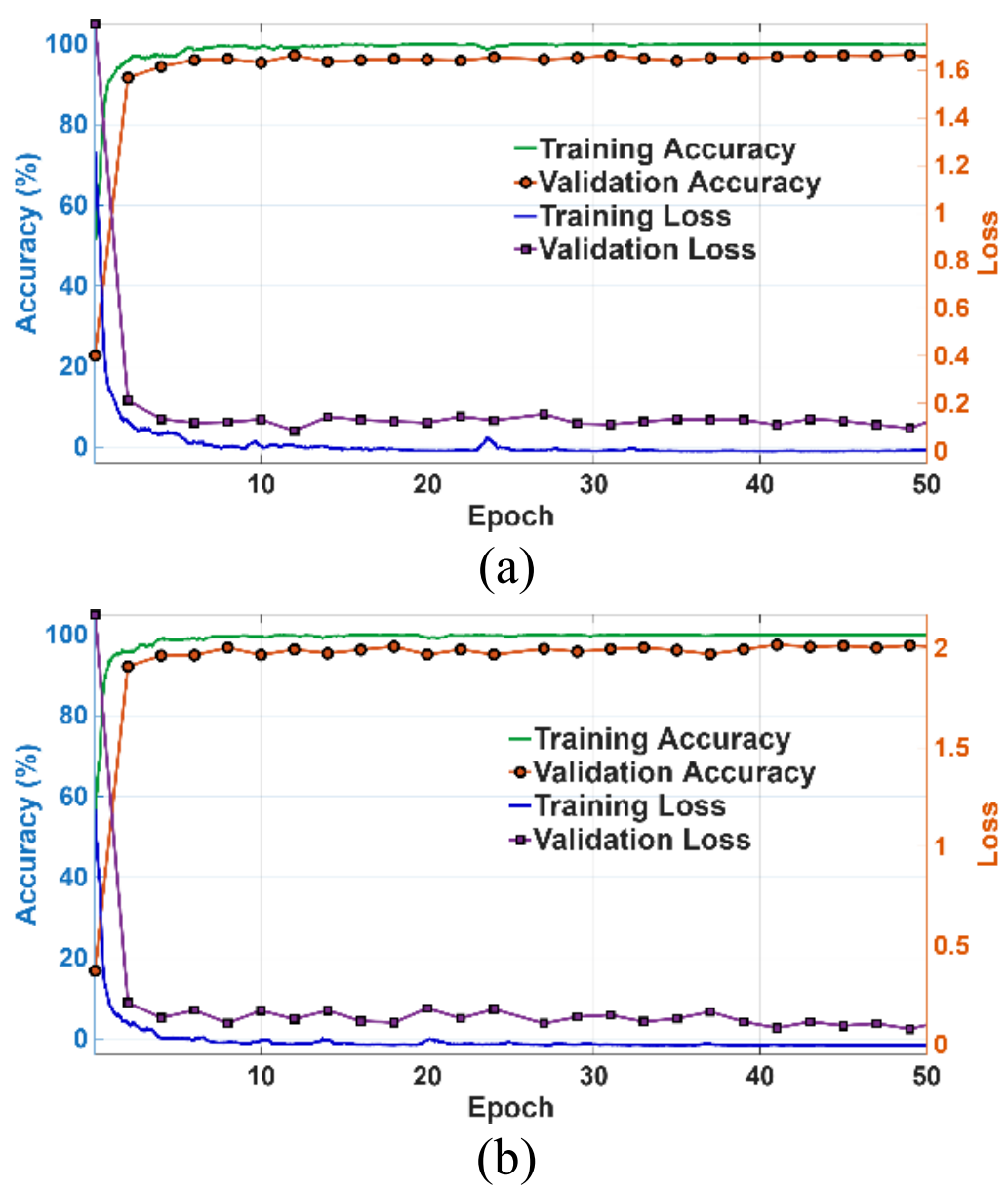

Fig. 8. Training and validation accuracy and loss curves for (a) InceptionV3 and (b) ResNet-18.

In image-classification problems, CNNs use convolutional layers to extract local spatial patterns and progressively combine them into higher-level features for classification [14], [30]. Since the proposed AWA representation is generated as an image-like pattern, CNN-based transfer learning was used in this work to evaluate whether class-dependent spatial distributions in AWA images can be learned directly. This allows classification without explicitly defining handcrafted features, unlike the Random Forest.

In this work, two pretrained CNN models based on the InceptionV3 and ResNet-18 architectures were selected to compare two established feature-learning mechanisms for image classification: multi-scale convolution in InceptionV3 and residual learning in ResNet-18. Fig. 7 presents both models within a common transfer-learning framework and shows the simplified feature-extraction structure of each network. InceptionV3 uses parallel convolutional filters with different receptive-field sizes to extract multi-scale image features [31], [32], while ResNet-18 uses residual connections to support effective feature learning in a relatively compact network structure [33]. For the PD source classification task, the pretrained convolutional layers of each network served as the feature-learning component, and the original classification layers were replaced with a new six-class classification head corresponding to the six PD classes considered in this work: C, I, S, CI, CS, and SI. The new classification head consisted of a fully connected layer, a softmax layer, and a classification layer for six-class PD source prediction. The AWA images were resized according to the input requirements of each network before training; InceptionV3 used images resized to 299 × 299 × 3, while ResNet-18 used images resized to 224 × 224 × 3.

During the training process, the labeled AWA images were supplied to the modified networks using the training dataset, while the validation subset was used to monitor model performance during learning. A cross-entropy loss function was used for multi-class classification, and the network parameters were updated using stochastic gradient descent with momentum (SGDM). The initial learning rate, mini-batch size, and maximum number of epochs were set to 0.001, 32, and 100, respectively. Training was performed using GPU acceleration, and validation accuracy was monitored to assess convergence and potential overfitting. After training, the final model performance was evaluated using the independent testing subset. All CNN models were trained on a workstation equipped with an NVIDIA GeForce RTX 4070 laptop GPU with 8 GB VRAM, an Intel Core Ultra 9 processor, and 16 GB system memory. GPU acceleration was used during training to improve computational efficiency, and the same hardware configuration was maintained for both CNN models to support fair comparison of training and testing time.

## V. Results and Discussion

TABLE I

Performance Comparison of Random Forest, InceptionV3, and ResNet-18

| Model | Training time per Epoch (s) | Testing Time per Image (ms) | Testing Accuracy (%) |
|---|---|---|---|
| **Random Forest** | – | 0.214 | 73.33 |
| **InceptionV3** | 99.09 | 1.742 | 96.53 |
| **ResNet-18** | 20.89 | 0.422 | 96.47 |

### A. Quantitative Performance Evaluation

The quantitative performance of the proposed PD classification framework was evaluated using three models: Random Forest, InceptionV3, and ResNet-18. The evaluation focused on testing accuracy and computational performance for the six PD classes. The overall results are summarized in Table I, where the training time per epoch is reported only for the CNN models because epoch-based training is not applicable to the Random Forest classifier.

The Random Forest classifier achieved a testing accuracy of 73.33%. Although it required a short model training time of 3.38 s and a testing time of 0.214 ms per image, its classification performance was lower than that of the CNN-based models. This result indicates that handcrafted features may not fully represent the complex spatial distributions present in AWA patterns for multi-class PD classification. In contrast, both CNN-based models achieved higher classification accuracy. The testing accuracies of InceptionV3 and ResNet-18 are 96.53% and 96.47%, respectively. Both models achieved the same validation accuracy of 97.93%, indicating effective learning of discriminative features from the AWA images. These results suggest that direct image-based feature learning is more effective for this dataset in the proposed AWA framework.

From a computational perspective, the results show that model architecture affects processing time, with ResNet-18 achieving accuracy comparable to InceptionV3 at lower computational cost. InceptionV3 required the longest training time per epoch, approximately 99.09 s, reflecting its higher computational complexity. ResNet-18 reduced the training time per epoch to approximately 20.89 s. Similarly, the testing time per image for ResNet-18 was 0.422 ms, which was lower than that of InceptionV3 at 1.742 ms, supporting a more computationally efficient implementation. The training behavior of the CNN models is illustrated in Fig. 8. Both models exhibit rapid convergence within the initial training epochs, followed by stable performance with high validation accuracy and low validation loss. The close agreement between training and validation accuracy indicates stable learning behavior without strong evidence of severe overfitting. Overall, the results indicate that CNN-based models outperform the Random Forest classifier in terms of classification accuracy while maintaining practical computational performance. Among the CNN models, ResNet-18 provides a favorable balance between accuracy and efficiency whereas InceptionV3 offers marginally higher accuracy at the expense of increased computational cost.

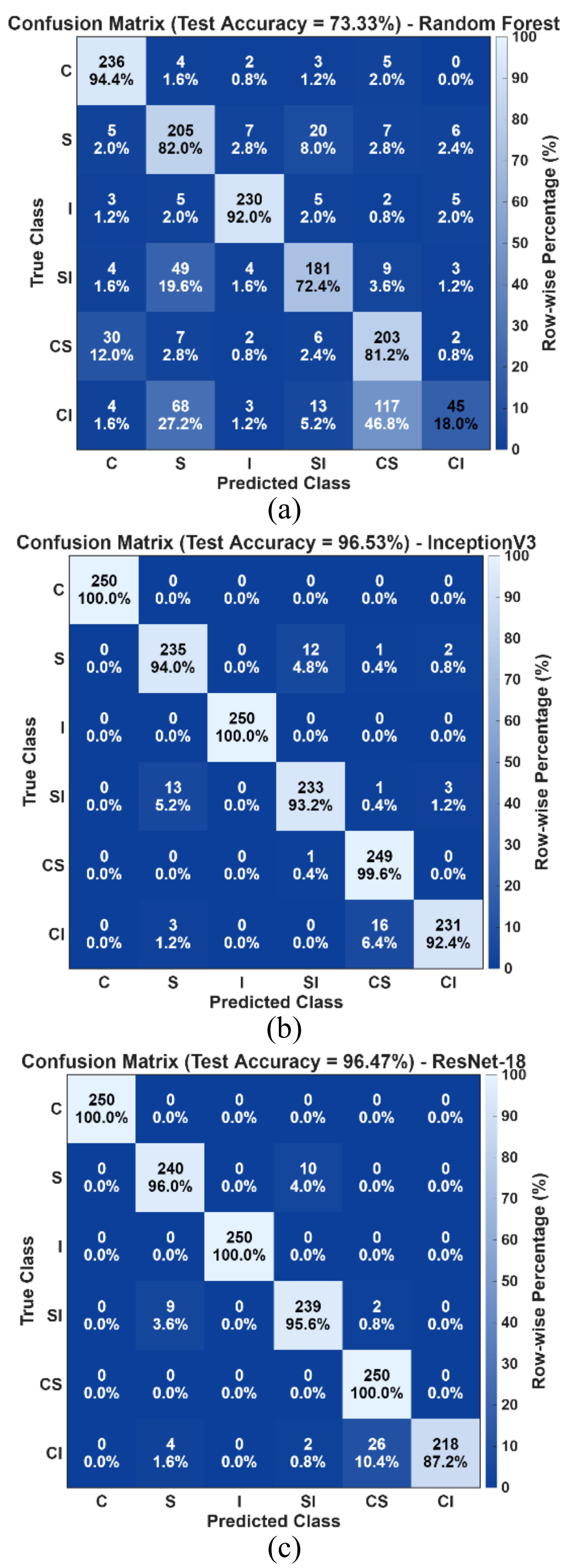

Fig. 9. Confusion matrices for (a) Random Forest, (b) InceptionV3, and (c) ResNet-18.

### *B. Classification Behavior and Model Analysis*

The class-wise classification behavior of the three models was analyzed using the confusion matrices shown in Fig. 9. Each confusion matrix is row-normalized, where the rows represent the true PD classes and the columns represent the predicted classes. Therefore, the diagonal elements indicate the percentage of correctly classified source of partial discharge and the off-diagonal elements indicate misclassification among different sources.

The Random Forest showed high accuracy for some single-source PD classes, particularly corona and internal PD, with accuracies of 94.4% and 92.0%, respectively. However, lower accuracy was observed for mixed PD conditions, especially for the CI class. A considerable portion of CI samples were classified as CS and S, indicating difficulty in separating mixed-source patterns using handcrafted features alone. This behavior suggests that the manually extracted spatial, color, and grid- based features provided limited separation for overlapping pattern characteristics in mixed PD classes.

In contrast, the CNN-based models showed stronger diagonal dominance across the confusion matrices, indicating more consistent classification performance. Both InceptionV3 and ResNet-18 correctly classified C and I PD with 100% class-wise accuracy. The CNN models also improved the classification of surface and mixed PD conditions compared with Random Forest. For InceptionV3, the main misclassifications occurred between SI and S, and between CI and CS. A similar trend was observed for ResNet-18 where the largest remaining confusion was between CI and CS. These misclassifications are reasonable because mixed PD patterns contain partially overlapping source-related features in the AWA representation. Overall, the confusion matrix analysis indicates that the AWA patterns contain class-discriminative information for both single and mixed PD sources, with InceptionV3 and ResNet-18 showing similar class-wise behavior and more effective learning of the spatial distribution characteristics of AWA images than the Random Forest.

## VI. Conclusion

This work presented an AWA–CNN framework for source-oriented classification of single and mixed partial discharges under switching-voltage excitation. Time-domain PD pulses were characterized using amplitude, width, and area, and these pulse-level features were mapped into AWA image patterns. The resulting representation produced distinct distributions for corona, internal, surface, corona+internal, corona+surface, and internal+surface PD conditions. The classification results showed that CNN-based image learning was more effective than handcrafted feature-based Random Forest classification for the proposed AWA patterns. ResNet-18 provided a favorable balance between accuracy and computational efficiency, achieving comparable accuracy to InceptionV3 with lower training and testing time. The confusion matrix analysis further indicated that CNN-based models provided higher class-wise recognition, especially for mixed PD conditions where overlapping pattern characteristics were present. Overall, the results suggest that the proposed AWA representation, combined with CNN-based transfer learning, provides an effective approach for multi-class PD source classification under switching-voltage excitation. Future work will focus on expanding the dataset across additional insulation materials, voltage waveforms, and operating conditions to further evaluate the generalization capability of the proposed framework.